# A Novel Way of Identifying Cyber Predators

Dan Liu
CENPARMI, CS & SE Deptartment
Concordia University
Montreal, Canada
l_dan17@encs.concordia.ca

Ching Yee Suen
CENPARMI, CS & SE Deptartment
Concordia University
Montreal, Canada
suen@cs.concordia.ca

Olga Ormandjieva
CENPARMI, CS & SE Deptartment
Concordia University
Montreal, Canada
ormandj@cse.concordia.ca

*Abstract*—Recurrent Neural Networks with Long Short-Term Memory cell (LSTM-RNN) have impressive ability in sequence data processing, particularly for language model building and text classification. This research proposes the combination of sentiment analysis, new approach of sentence vectors and LSTM-RNN as a novel way for Sexual Predator Identification (SPI). LSTM-RNN language model is applied to generate sentence vectors which are the last hidden states in the language model. Sentence vectors are fed into another LSTM-RNN classifier, so as to capture suspicious conversations. Hidden state enables to generate vectors for sentences never seen before. Fasttext is used to filter the contents of conversations and generate a sentiment score so as to identify potential predators. The experiment achieves a record-breaking accuracy and precision of 100% with recall of 81.10%, exceeding the top-ranked result in the SPI competition.

## I. INTRODUCTION

Greater popularity of social networks gives rise to cyber-criminal activities conducted by Sexual Predators (SPs). In this context, PAN initiated the Sexual Predator Identification (SPI) Task in 2012 [1]. PAN collects and shares an overwhelming amount of data on online chats, inside which there are predators, to facilitate research in predator behaviors.

There are two separate tasks in SPI, namely, identification of SPs inside chats and highlighting specific SPs lines in chats. The research reported in this paper focuses on the first task. As [1] indicated, the first step is to find out which conversations are suspicious, then identify such conversations to belong to which author, [2], [3], and [4] use a similar method to identify the predators.

However, the organizer set the goal for SPI task as creating a large and realistic dataset. The side effects of realistic data is high noise level, unbalanced training samples, and various lengths of conversations (from 1 to more than 500). More specifically, there are lots of general and sex-related conversations, while among them only a few involve SPs. Furthermore, there are a lot of chat abbreviations and cyber slangs in conversations, such as "ur" for "your", "yr" for "year", "sorryyyy" for "sorry", to name a few. Such words are crucial to feature selection and should be considered in the process of feature selection. Therefore, traditional machine learning methods cannot achieve satisfying performance unless with data truncation using numerous rules. Even if n-gram is used, with hundreds of thousands of conversations, the noise will result in extreme sparsity, and the performance will be weakened consequently [5].

LSTM-RNN [6] sentence vectors are introduced to solve the above-mentioned noise and performance problems. Different from n-gram, sentence vectors are able to capture sentence features more efficiently and compress the size of input data, as the classifier will only take sentences, instead of words, as features. Meanwhile, LSTM-RNN classifier can also be used for suspicious conversation detection (SCD) as it is good to learn long-term dependencies in time series data. The experiment generated an accuracy rate of 99.43% on SCD and 98.35% on SPI, respectively. Finally, 206 out of 254 predators were identified by the intersection of two classifiers with zero error, which exceeded the best result [2] of the official ranking (203 out of 254 with 3 misclassifications).

The contributions of this paper are three-fold, namely: i) LSTM-RNN is introduced to generate sentence vectors especially for sentences never seen before with known words; ii) IMDB reviews [7] is used to test the performance of sentence vectors model, and iii) ssentiment score is introduced to improve the sexual predators identification performance.

The rest of the paper is organized as follows: the related work is summarized in section II. Our approach is explained in section III. Section IV describes the experimental work carried on in this research. Finally, section V, VI concludes the paper.

## II. RELATED WORK

A common strategy for SPI is the use of two classifiers. The first classifier will detect suspicious conversations which can be seen as positive (with predators) or negative (no predators involved) [2, 3, 4]. However, very complex and specific rules were applied to remove noise or to extract features. Especially, in [2], only about 10% samples remained for training and testing. Such removal could influence the generalization ability of the classifier. Manual rules for features extraction in [8, 9, 10] will reduce the stability as only samples that match the rules can be classified. Neural network language model approach can conquer those problems mentioned above. The second classifier is about predator identification. Support Vector Machines (SVM) [8, 11], Naive Bayes [12, 13] and other classical machine learning approaches [3, 5, 9, 14] were introduced. From official rank [1], those approaches whose precision was greater than 90% had a lower recall (less than 80%).

Neural network language model is introduced in [15]. After that a series of derived version [16, 17, 18] ,i.e. word embedding, sentence embedding etc. is widely applied in Natural Language Processing (NLP) work. For the sentence embedding or document embedding, it shows very strong ability in NLP tasks. Because it is hard for those language models represent sentences never seen before, the new sentence which is not in the training dataset will be a problem. Unless those embeddings, [19] proposed a more compatible method with sequence to sequence model to vectorize sentences based on previous and next sentences.

There are two mainly types of sentiment analysis task in NLP area online service rating [20] and movie review [7]. Severyn and Moschitti [21] used the distance to margin of SVM as sentiment score, the larger it is the more positive or negative it will be. Deep neural network is also a popular area for sentiment analysis. Reference [22] compared the performance of different neural network models, such as NBSVM-bi [23] and LSTM-RNN etc. Although the deep neural networks have strong capability on NLP tasks, the training cost of deep neural network cannot be neglected. Fasttext [24], is a neural network with shallow and beautiful structure which had excellent performance on sentiment analysis and language model tasks.

## III. APPROACH

The experiment involves three types of neural networks. Firstly, the LSTM-RNN-based language model, which is used to express the relation inside a sentence. The last hidden state of LSTM-RNN of each sentence will be used as sentence vectors. Secondly, a two-layer LSTM-RNN classifier is used to find suspicious conversations by learning the dependencies among sentences in a conversation. Each sentence in a conversation will be regarded as a single timestep and fed into the classifier. Lastly, following the detection of suspicious conversations, a Fasttext-based sentiment classifier is introduced to identify SPs. The conversations are split into different groups by author. In this way, the SPs can be identified sentiment score.

### A. Processing

PAN2012 dataset contains a great number of chat abbreviations, cyber slangs, emoticons and conversations of various lengths, which will increase the perplexity of language model. To predict the next word more accurately, noise removal and replacement are indispensable. However, some words or abbreviations may convey important information, for example, yrs means years, ur means your, etc., therefore recovery of these abbreviations is necessary. Those strategies are listed below:

- Replace the number by symbol 00NUM.
- Replace the words longer than 30 characters by 00LW.
- Replace the URL with symbol 00URL in the data.
- Remove all non-ascii chars.
- Remove all emoticons.
- Recover popular yet unofficial abbreviations.

### B. Recurrent Neural Network

Recurrent neural networks (RNNs) are neural network models that process elements of a sequence one by one and learn the dependencies among previous inputs. There are three layers, i.e. Input X, Hidden S, and Output Y. The input of RNN at time step t is $x_t \in R^n$ and the hidden state is $s_t \in R^m$.

$$s_t = f(Wx_t + Us_{t-1}) \quad (1)$$

$s_t$ is calculated by (1), based on the previous hidden state $s_{t-1}$ and the current input step $x_t$ where the function $f$ is a nonlinear function. The output is (2).

$$y_t = Softmax(Vs_t) \quad (2)$$

RNN is featured by its ability to capture dependencies in sequences and share the same parameters $(U, V, W)$ throughout all steps. Theoretically, RNN learns through all previous timestep, however, due to vanishing gradients problem [25], it is hard to capture long-term dependencies.

LSTM-RNN proposes a gating mechanism to avoid vanishing gradients problem. More specifically, a new state $c_t$ is introduced to calculate hidden state $s_t$ (Fig. 1). $c_t$ and $s_t$ are calculated as below:

$$i_t = \sigma(x_t U^i + s_{t-1} W^i) \quad (3)$$
$$f_t = \sigma(x_t U^f + s_{t-1} W^f) \quad (4)$$
$$o_t = \sigma(x_t U^o + s_{t-1} W^o) \quad (5)$$
$$g_t = tanh(x_t U^g + s_{t-1} W^g) \quad (6)$$
$$c_t = f_t * c_{t-1} + i_t * g_t \quad (7)$$
$$s_t = o_t * tanh(c_t) \quad (8)$$

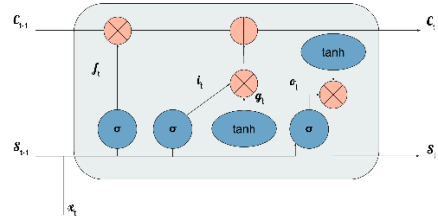

Fig. 1. LSTM-RNN cell.

where $i_t$, $f_t$ and $o_t$ are input, forget gate and output gate. $g_t$ is the hidden state like $s_t$ in RNN which is used to compute new $s_t$.

### C. Recurrent neural network language model

The neural network language model takes a word sequence $W = [w_1, ..., w_t], w_t \in V$ where $V$ is the vocabulary set as input and learns to predict the probability $p(w_{t+1}|w_1, ..., w_t)$ of the next word $w_{t+1}$ by applying Softmax activation function at the output layer. $P(w_{t+1}|w_1, ..., w_t) = Softmax(s_t^T e_{wt})$ where $e \in E[e_{w1}, .... e_{wt}]$ [15]. This input is mapped to vectors $e_{wt}$ in feature space within the neural network. The training of neural network uses error back-propagation algorithm over time to maximize the log-likelihood (9) of training data.

$$L(\theta) = \sum_t \log P(w_t | w_{t-n+1}, ... w_{t-1}) \quad (9)$$

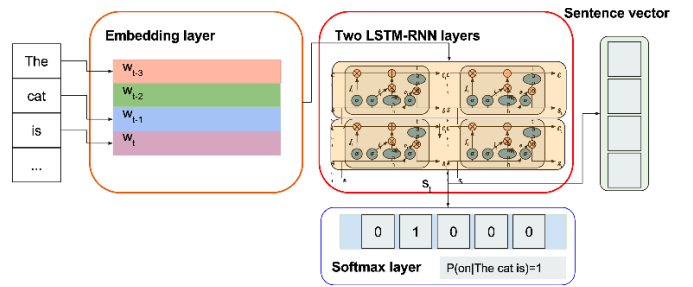

Fig. 2. LSTM-RNN language model.

## D. Sentence vectors

LSTM-RNN neural network language model with two hidden layers is used to generate sentence vectors (Fig. 2). Specifically, the last time step hidden state in LSTM-RNN language model $s_t$ is used to represent the input sequence $\{w_1, \ldots, w_t\}$. The first layer of this model is the embedding layer that represents words in dense vectors. The second and third layers are LSTM-RNN that learn the dependencies among the words in sentences. The final layer is the Softmax layer, a multinomial logistic regression layer used to solve multi-class prediction problems. Compared with word embedding, the sentence vectors presentation can reduce the length of inputs. LSTM-RNN-based sentence vectors have the advantages of being able to capture the dependency and compress the size of conversations. In the meantime, those words with term-frequency of less than 10 are removed as noise and the remaining words are sorted by term frequency–inverse document frequency (TF-IDF) weights.

## E. Conversation Classification

With regard to suspicious conversation identification, LSTM-RNN has strong ability to learn the long-term dependencies among timesteps, which means it can capture relations among sentences that contain the features of predators. Conversations with sentence vectors are input into a three-layer LSTM-*RNN*-based model and the latter will learn the context features (Fig. 3). Considering different number of sentences in conversations (from 1 to more than 500), those extra-long conversations will be padded by zeros and then split into parts, each with an equal length of 100 (an experience-based value). This strategy will prevent underfitting in LSTM-RNN model when processing long conversations as there are only a few of them. Due to the well-distributed features in suspicious conversation, a predator is very likely to carry out criminal activities throughout an entire conversation.

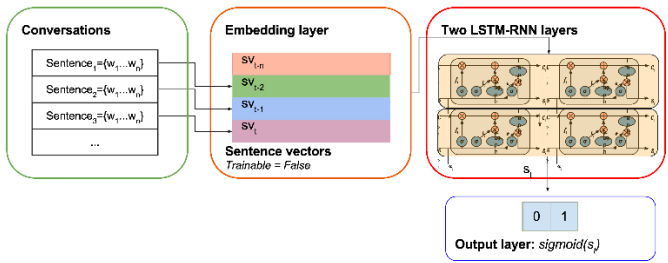

Fig. 3. LSTM-RNN classifier.

## F. Author Classification

Different from LSTM-RNN, which is good at capturing time series features, Fasttest is a very shallow neural network capable of global feature extraction. Once the suspicious conversations are identified by LSTM-RNN model, the Fasttext-based classifier can identify SPs among the authors. In addition, those conversations only containing predators will not be deleted because there might be useful features inside such conversations. In order to improve the accuracy of SPs identification, sentiment score is assigned. There are three types of scores, i.e. P, V, N, to differentiate authors from predators, victims, and normal users. The output of Softmax layer is the most ideal score model. An author with a higher score in a scoring type among the three will be classified into that group. It is unlikely for two predators to appear in the same conversation, therefore an author with the highest score in P category will be identified as predator. As there might be conversations initiated by the same author at different time, the sentiment score of those authors is averaged.

## IV. EXPERIMENTS

For the purpose of detecting SPs, firstly, PAN2012 dataset is used to train the neural network language model and the SCD classifier. After that, Fasttext classifier is trained on PAN2012 dataset grouping by authors. The IMDB sentiment review dataset is used to evaluate the performance of the model, i.e., sentence vectors methodology.

### A. Dataset

The performance of the models is evaluated on two datasets, PAN2012 dataset and Stanford Large Movie Review Dataset (IMDB sentiment review dataset) [7]. Performance will be measured against existing publications on sentiment classification tasks.

*1) PAN2012 Dataset*

PAN2012 dataset is provided in the context of Sexual Predator Identification (SPI) Task in 2012 initiated by PAN (Plagiarism analysis, authorship identification, and near-duplicate detection) lab. In the training dataset, there are 66,927 chat conversations with over 97,000 different users and only 142 of them are SPs. The test dataset contains 155,128 chat conversations with over 218,000 different users and only 254 of them are SPs (TABLE I).

TABLE I. ATTRIBUTES OF PAN2012 DATASET

|  | Training | | Test | |
| --- | --- | --- | --- | --- |
| *Type* | *Original* | *Filtered* | *Original* | *Filtered* |
| Positive | 2,016 | 1,088 | 3,684 | 1,880 |
| Negative | 64,911 | 52,854 | 151,210 | 123,229 |
| Non-predators | 97,547 | 97,291 | 218,488 | 217,997 |
| Predators | 142 | 138 | 254 | 215 |

TABLE II. SENTENCE LENGTH DISTRIBUTION OF PAN2012

|  | Training dataset | | Test dataset | |
| --- | --- | --- | --- | --- |
| *Number of words* | *Positive* | *Negative* | *Positive* | *Negative* |
| 0-20 | 1,142 | 51,362 | 2,189 | 119,545 |
| 21-40 | 182 | 6,245 | 274 | 14,711 |
| 41-60 | 129 | 2,482 | 243 | 5,971 |
| 61-80 | 113 | 1,326 | 255 | 3,163 |
| 81-100 | 110 | 853 | 217 | 2,003 |
| >100 | 291 | 2,576 | 506 | 5,817 |

TABLE III. EXPERIMENTAL DATA OF NO.1 IN PAN2012 COMPETITION (*VILLATORO-TELLO ET AL. 2012*).

| Number of... | Original data | Filtered data |
|---|---|---|
| Chat conversations | 66,928 | 6,588 |
| Users | 97,690 | 11,038 |
| Sexual Predators | 148 | 136 |

*2) IMDB movie reviews*

IMDB Large Movie Review Dataset provides 50,000 binary labeled reviews extracted from IMDB. In this dataset, highly polar movie reviews, with a score lower than 4 or higher than 7 on a scale of 10, is split evenly into 25,000 training samples and 25,000 test samples. The overall distribution of labels is balanced. The distribution of sentence length of reviews is shown below (TABLE IV).

TABLE IV. SENTENCE LENGTH DISTRIBUTION OF IMDB DATASET

| | Training dataset | | Test dataset | |
|---|---|---|---|---|
| *Number of words* | *Positive* | *Negative* | *Positive* | *Negative* |
| 0-20 | 6,807 | 6,943 | 7,051 | 6,942 |
| 21-40 | 3,832 | 3,941 | 3,775 | 3,978 |
| 41-60 | 1,133 | 1,032 | ,1043 | 1,028 |
| 61-80 | 450 | 368 | 371 | 361 |
| 81-100 | 193 | 135 | 172 | 132 |
| >100 | 85 | 81 | 88 | 59 |

B. *Experimental setup*

*1) Suspicious conversation detection*

For the SCD task, LSTM-RNN language model is trained with the architecture shown in (Fig. 3). The sentence vectors are the last hidden state of LSTM-RNN language model. Each conversation being represented by a group of sentence vectors is fed into a new LSTM-RNN binary classifier. The performance of this classifier is demonstrated in (Fig. 4). In the methodologies mentioned above, one embedding layer, two LSTM-RNN layers with 200 units and 50 timesteps as well as a Softmax layer are implemented on Tensorflow framework for language model. The SCD classifier is implemented on Keras framework has the similar structure as LSTM-RNN language model, except that SCD classifier replaces Softmax layer with sigmoid layer.

*2) IMDB sentiment task*

The IMDB sentiment task is introduced for evaluating the classification performance of sentence vectors. This task shares the same structure and configuration as SCD task.

Related performance data is shown in Fig. 6, TABLE XI, and TABLE XII.

V. RESULTS

A. *Language model generated from PAN2012*

By using a LSTM-RNN neural language model with two hidden layers, together with 35 timesteps and 200 hidden units, the perplexity of language model on test dataset reached 10.948 after 30 iterations.

B. *Suspicious Conversation Detection*

The result of suspicious conversation detection task is shown in Fig. 4 and TABLE V. It is obvious that the best result is obtained at epoch 5. The accuracy of sentence-vector model is 99.43%, exceeding the accuracy of 98.83% with SVM obtained by [2] (TABLE VI).

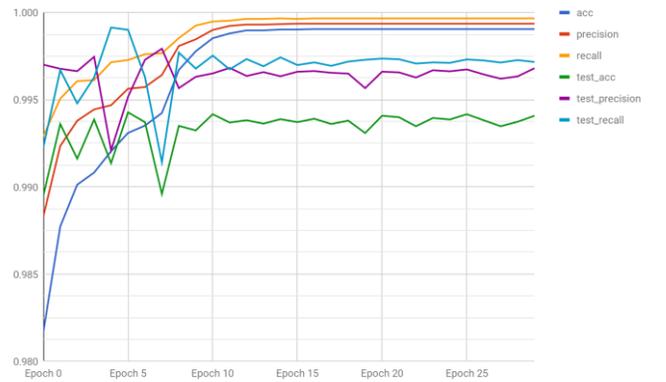

Fig. 4. The performance of SCD classifier.

TABLE V. BEST RESULT AT EPOCH 5.

| | Accuracy | Precision | Recall | F-1 Score |
|---|---|---|---|---|
| Training | 0.9931 | 0.9956 | 0.9973 | 0.9965 |
| Test | **0.9943** | 0.9955 | 0.9987 | 0.9971 |

TABLE VI. PERFORMANCE OF NO.1 IN PAN2012 COMPETITION (VILLATORO-TELLO ET AL. 2012).

| Algorithm | Weighting | Accuracy | F-measure |
|---|---|---|---|
| SVM | binary | 0.9848 | 0.9361 |
| SVM | tf-idf | **0.9883** | **0.9516** |
| NN | binary | 0.9874 | 0.9464 |
| NN | tf-idf | 0.9825 | 0.9254 |

C. *Sexual Predator Identification*

In the sexual predator identification task, the performance of Fasttext model is very stable. The training procedure is completed in 20 minutes. The sentiment score is generated from Softmax layer of the Fasttext model. The score of the same participants in different conversations is averaged. The result (TABLE X) shows that all predators have a very high sentiment score compared with victims.

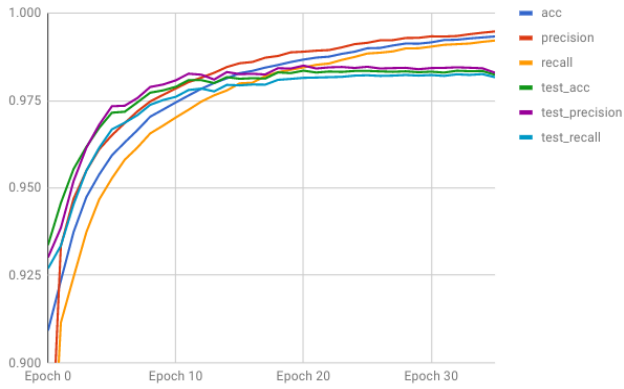

Fig. 5. The performance of SPI classifier.

TABLE VII. THE BEST RESULT OF SPI CLASSIFIER.

|  | Accuracy | Precision | Recall | F-1 Score |
|---|---|---|---|---|
| Training | 0.9899 | 0.9915 | 0.9885 | 0.9909 |
| Test | 0.9835 | 0.9847 | 0.9822 | 0.9835 |

TABLE VIII. THE RESULT AFTER APPLIED SENTIMENT SCORE.

| Retrieved documents | Relevant documents | Accuracy | Precision | Recall | F-1 Score | F-0.5 Score |
|---|---|---|---|---|---|---|
| 206 | 206 | 100.00% | 100.00% | 81.10% | 89.56% | 95.55% |

TABLE IX. THE OFFICIAL RANK RELEASED BY PAN LAB (G. INCHES, F. CRESTANI 2012).

Results for problem 1): identify predators. The table reports the evaluation of all the runs submitted ordered by value of F score with β = 0.5.
Runs with ranking number are the ones used for official evaluation.
RET. = Retrieved documents, REL. = Relevant document retrieved.
P = Precision. R = Recall

| Participant run | RETR. | REL. | P | R | $F_{\beta=1}$ | $F_{\beta=0.5}$ | rank |
|---|---|---|---|---|---|---|---|
| villatorotello | 204 | 200 | 0.9804 | 0.7874 | 0.8734 | 0.9346 | 1 |
| snider12 | 186 | 183 | 0.9839 | 0.7205 | 0.8318 | 0.9168 | 2 |
| villatorotello | 211 | 200 | 0.9479 | 0.7874 | 0.8602 | 0.9107 |  |
| parapar12 | 181 | 170 | 0.9392 | 0.6693 | 0.7816 | 0.8691 | 3 |
| morris12 | 159 | 154 | 0.9686 | 0.6063 | 0.7458 | 0.8652 | 4 |

TABLE X. PART OF THE PREDATORS AND VICTIMS WITH SENTIMENT SCORE.

| Predators | Score |
|---|---|
| 004ed4354a09e2c33117335adb24e333 | 0.97 |
| 00851429b21722a4d62f63a328c601ca | 0.99 |
| 00d36f64d208c95eeb70af477dfb368a | 1.00 |
| 00fe41de80eb7527c81f7915ab5a6479 | 0.67 |
| 013dab612d37dc4e2cce87da5239f537 | 0.92 |
| **Victims** | **Score** |
| 9eb10acea3e6eb0da7b37acef57a5097 | 0.03 |
| ef8fbb24e05c1d18efc7a75a812da6ed | 0.02 |
| 980ffbae20a666d965bb171413352750 | 0.01 |
| 001744005608bb20b997db6d8cabb3a9 | 0.27 |
| b3d822f188649acd6401e8289193184a | 0.02 |

*D. IMDB sentiment reviews*

*1) Language model generated from IMDB reviews*

The language model is built with the same method as SCD and the test perplexity is 126.903 which is not so good. The perplexity is significantly higher than SCD's. The timestep of this model is 50, which is longer than SCD's as the average number of sentence per input of IMDB dataset is larger (see TABLE II and TABLE IV).

*2) Sentiment results of IMDB*

Compared to the training result, the test result of the sentence-vector model on IMDB dataset indicates that the performance of this model is very unstable. Although the accuracy of 83.2% with sentence-vector model, the sentence-vector reduces the length of the inputs and accelerates the training and test speed.

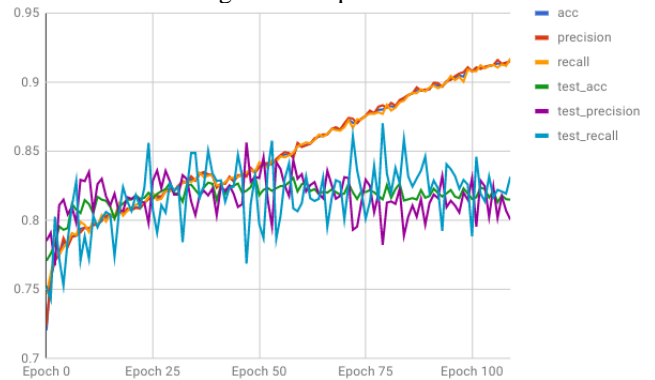

Fig. 6. The training and test performance on IMDB dataset.

TABLE XI. BEST RESULT AT EPOCH 49.

|  | Accuracy | Precision | Recall |
|---|---|---|---|
| Training | 83.43% | 83.46% | 83.49% |
| Test | 83.23% | 83.13% | 83.13% |

TABLE XII. COMPARISION ON THE IMDB SENTIMENT TASK.

| Model | Train | Test |
|---|---|---|
| NBSVM-bi (Wang and Manning, 2012) | N/A | 91.2% |
| Paragraph Vector (Le and Mikolov, 2014) | N/A | 92.7% |
| Paragraph Vector (Hong and Fang, 2015) | 97.1% | 94.5% |
| Sentence Vector + 2-layers-LSTM-RNN | 83.4% | **83.2%** |

VI. CONCLUSIONS

This paper presents a combined method to identify sexual predators. The sentiment score from Softmax layer outputs is crucial in the final identification step. The approach of taking LSTM-RNN last hidden state as sentence vectors is highly efficiency as the long conversations are shortened by sentence vectors. The reason of sentence-vectors-based classifier does not work well on IMDB dataset could be that the perplexity of its language model is too high to represent the sentence. In the future work, other language models will be applied to reduce the perplexity to see if the accuracy will be enhanced. In the meantime, attention mechanism [26] in neural network can also be introduced to detect keywords in predators' conversations.

ACKNOWLEDGMENT

We would like to thank Natural Sciences and Engineering Research Council of Canada for supporting this project.